\documentclass[sigconf]{acmart}
\usepackage{amsmath}
\usepackage{graphics}

\AtBeginDocument{%
  \providecommand\BibTeX{{%
    \normalfont B\kern-0.5em{\scshape i\kern-0.25em b}\kern-0.8em\TeX}}}

\setcopyright{acmcopyright}
\copyrightyear{2020}
\acmYear{2020}

\setcopyright{none}
\begin{document}

\title{Reference Based Color Transfer for Medical Volume Rendering}

\author{Sudarshan Devkota}
\email{sudarshan.devkota93@knights.ucf.edu}
\affiliation{%
  \institution{University of Central Florida}
  \streetaddress{4000 Central Florida Blvd}
  \city{Orlando}
  \state{Florida}
  \postcode{32816}
}

\author{Sumanta Pattanaik}
\email{sumant@cs.ucf.edu}
\affiliation{%
  \institution{University of Central Florida}
  \streetaddress{4000 Central Florida Blvd}
  \city{Orlando}
  \state{Florida}
  \postcode{32816}
}

\renewcommand{\shortauthors}{Devkota, et al.}

\begin{abstract}
The benefits of medical imaging are enormous. Medical images provide considerable amounts of anatomical information and this facilitates medical practitioners in performing effective disease diagnosis and deciding upon the best course of medical treatment. A transition from traditional monochromatic medical images like CT scans, X-Rays or MRI images to a colored 3D representation of the anatomical structure further enhances the capabilities of medical professionals in extracting valuable medical information. The proposed framework in our research starts with performing color transfer by finding deep semantic correspondence between two medical images: a colored reference image, and a monochromatic CT scan or an MRI image. We extend this idea of reference based colorization technique to perform colored volume rendering from a stack of grayscale medical images. Furthermore, we also propose to use an effective reference image recommendation system to aid for selection of good reference images. With our approach, we successfully perform colored medical volume visualization and essentially eliminate the painstaking process of user interaction with a transfer function to obtain color and opacity parameters for volume rendering.  
\end{abstract}

\ccsdesc[300]{Human-centered computing~Scientific visualization}

\keywords{Deep Learning, color transfer, volume rendering, medical imaging}

\begin{teaserfigure}
  \includegraphics[width=\textwidth]{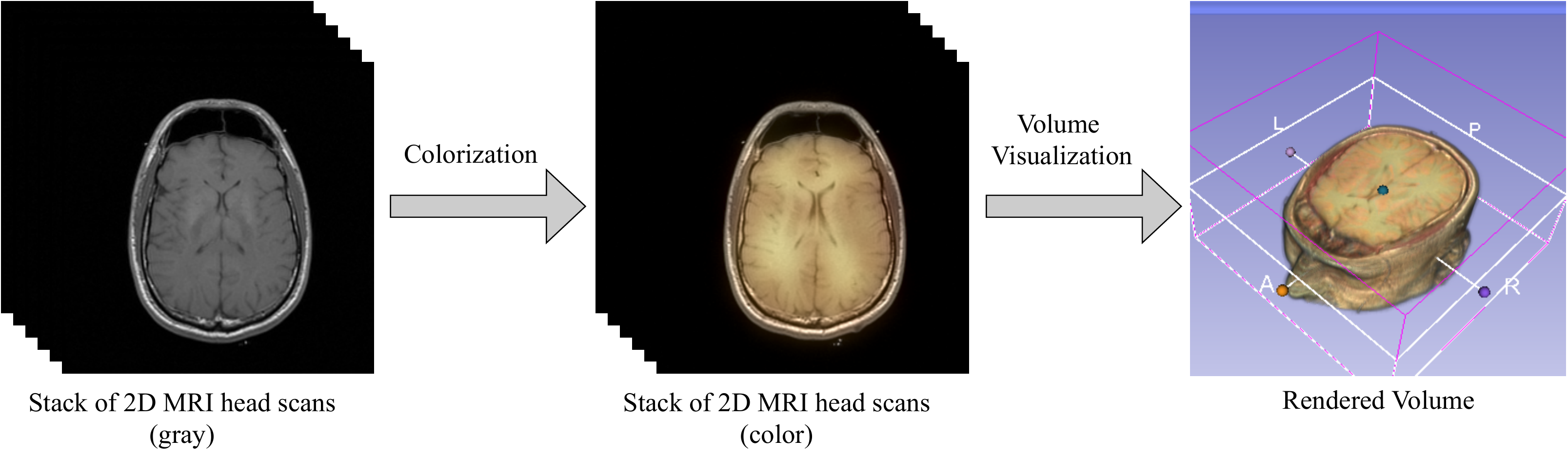}
  \caption{ The proposed system comprises of colorization of grayscale medical images followed by direct volume rendering. }
  \label{fig:fig1}
\end{teaserfigure}

\maketitle

\section{Introduction}
Medical imaging is an extremely important aspect in the medical field and has revolutionized the health care sectors, allowing for earlier diagnoses of diseases and better patient outcomes. It has brought a significant change in the healthcare sector and has aided scientists and medical practitioners to learn more about the human body than ever before. Medical images like X-rays, CT scans and MRIs are a valuable diagnostic tools and contain a significant amount of anatomical information for clinical procedures. However, these images are generally gray-scale images and a colored image could provide better anatomical information than an ordinary monochromatic image.

The idea of colorizing grayscale images to make them appear aesthetically pleasing is not new. In the last decade, several techniques have been proposed for colorization of images and videos. While some of these techniques involve propagating colored scribbles taken from users to achieve the final colorized result \cite{Levin04}\cite{Huang05}\cite{Luan07}\cite{Qu06}, other learning based techniques leverage large-scale image dataset to achieve automatic and reliable colorization without any user effort \cite{Deshpande15} \cite{Cheng15} \cite{Zhang16}\cite{Larsson16}\cite{Iizuka16}. Another category of colorization technique is example based colorization that seeks to transfer color properties from a reference image to a target image \cite{Welsh02}\cite{Irony05}\cite{Gupta12}\cite{Bugeau14}. In an example based technique, establishment of semantic similarities between the images plays a crucial role in performing accurate color transfer. Due to large variations in structural content or appearance, establishing semantic similarities could be challenging with matching methods based on low level hand-crafted features like intensity patch features, Haar features, HOG features, SIFT features, etc. A recent work performed by Liao et al.\cite{Liao17} demonstrated that multi-level features obtained using deep neural networks are robust in finding high level semantic correspondence between images, thus reducing the limitation of using hand-crafted features.

In our work, we take an approach similar to the one used by Liao et al., leveraging a pretrained network like VGG-19\cite{VGG19} as the feature extractor network and using multi level features to establish deep correspondence between images. While most of their approach serves as a base to our implementation, we extend it to color transfer for medical data and medical volume rendering with direct volume rendering technique. Specifically, we perform colorization of monochromatic MRI images and CT scans with a specific set of reference images and perform medical volume visualization with the colored version of medical images. 

Medical Volume Visualization provides meaningful information from the rendered three-dimensional structure and it allows medical professionals like surgeons, physicians and radiologists to better visualize the anatomy and understand the disease processes. Furthermore, the shift from grayscale volume visualization, where each voxel constitutes a single intensity value, to a colored 3D volume, where each voxel has RGB components, further provides improved visualization of the anatomical structures. This also enables medical practitioners in reviewing large medical dataset in an effective and comprehensive way. 

To perform a colored 3D rendering, each voxel must have $RGBA$ value, where $RGB$ defines the color channels and $A$ is the opacity channel. Voxels that have a low alpha value will appear transparent while voxels with high alpha value will appear opaque. One essential step in traditional volume rendering technique is to design proper transfer functions, where a user interacts with a volume renderer by modifying values in the transfer function in order to obtain best color and opacity parameters in the final rendered volume. This work essentially involves trial-and-error interaction with the transfer function, and can be tedious and time consuming. With our proposed system, we obtain the color values with reference based colorization technique, and for the alpha channel, we use the luminance values of each voxel as the alpha value during volume rendering.

In brief, our major technical contributions are:
\begin{itemize}
  \item We experimentally verify the effectiveness of multi-level feature extraction to establish deep correspondence and to eventually perform color transfer between two medical images, where one is a colored reference image, and other is a CT scan or a MRI image.
  \item We extend color transfer with deep image analogy for medical volume visualization. Our approach essentially removes tedious user interaction with transfer function to obtain color parameters for volume rendering.
\end{itemize}

The overall layout of the paper is organized as follows. 
In section \ref{section:relatedWork}, we discuss some of the existing methods in the field of image colorization and volume rendering. In section \ref{section:dataset}, we explain about the image dataset that we use in our proposed system. In subsection \ref{section:meth-deepImageAnalogy}, we explain visual attribute transfer using a state of the art dense correspondence algorithm like Deep Image Analogy. This is followed by colorization using guided image filters in subsection \ref{section:meth-colorization}, where we elaborate on the further processing that is needed on the output of deep image analogy to preserve only color and disregard other visual attributes that are transferred from reference image to target monochromatic medical image. In section \ref{section:meth-referenceImageRetrieval}, we explain a reference image recommendation engine that automatically finds good reference images for color transfer. The recommendation algorithm uses deep-features obtained from a pre-trained network like VGG-19 to select a top few reference images which are semantically similar to the target grayscale image. Furthermore, we briefly explain the approach we’ve taken for volume rendering in section \ref{section:meth-DVR}. And finally in section \ref{section:results}, we present the medical image colorization and final volume rendering results.

\section{Related Work}
\label{section:relatedWork}

With the recent advances in neural networks, there have been several works to show that deep learning approaches can be successfully applied to various fields, and in some applications, there are cases where the results obtained using these networks have even been superior to what human experts can achieve. In this section we review the work that has been performed in the following two sectors: Color Transfer and Volume Rendering.

\subsection{Colorization}
\label{section:rw-colorization}

One of the novel techniques on color transfer which was proposed by Levin et al. required color scribbling by users on top of the grayscale images and then using an optimization technique for colorization of the input image \cite{Levin04}. There have been several advancements on this pioneering work. Qu et al. extended their idea for colorization of manga which contains different pattern and intensity continuous regions\cite{Qu06}. These work showed some promising results but one of the drawbacks of these methods is that they require proper scribbling with intense manual effort. 

In recent years, with capabilities of Deep Neural Networks, different learning based techniques have been proposed for automatic colorization, which do not require any user effort. Cheng et al. \cite{Cheng15} used a deep neural network to colorize an image where they pose the colorization problem as a regression problem. With comparatively larger set of data, Zhang et al. \cite{Zhang16}, Larsson et al. \cite{Larsson16} and Iizuka et al. \cite{Iizuka16} used Convolutional Neural Networks (CNNs) to perform automatic colorization, where the three methods differ in terms of loss functions and CNN architectures. The first two methods used classification loss, while the third one used regression loss. While all of these networks are learned from large-scale data, there have been other works \cite{Liao17} \cite{He17} that perform color transfer from a single colored reference image to a grayscale image. These methods utilize pretrained networks like VGG-19 for extracting deep features from the images. These deep features are used to find reliable correspondence between the images which is further leveraged to perform attribute transfer like style and color. While the work we perform is primarily based on Liao et al.'s \cite{Liao17} work on visual attribute transfer, we extend their idea to perform color transfer with various guided image filters, and focus our work towards medical dataset. We further extend our work for medical volume rendering.

\subsection{Volume Rendering}
\label{section:rw-volumeRendering}

In the last decade, research in volume rendering has expanded to a wide variety of applications. 
Conventional volume rendering pipeline implements either one dimensional or multidimensional transfer functions \cite{Kindlmann98} \cite{Kniss02}\cite{Roettger05} for volume rendering. 
Volume rendering with multidimensional transfer functions than 1D transfer functions have been immensely successful in producing better results, however, with the introduction of more features for classification of voxels, it makes interaction between users and the multidimensional transfer function more complex.

In  an  attempt  to  take  a  different  approach  to  volume  rendering,  a  certain number of works \cite{He96}\cite{Berger19}\cite{Hong19}  have utilized the capabilities of machine learning techniques by introducing a new pipeline for volume visualization. While these methods have been successful in producing exceptional results,  the  majority  of  them  require  a  significantly  large  amount  of  training data. In our method, we require a comparatively smaller set of reference images to obtain the color properties for volume rendering. Furthermore, our work only focuses on coloring part of 1D transfer function by automatic mapping of intensity to color without any user effort. Obtaining opacity parameters from a transfer function also plays a significant role in direct volume rendering, but we are not addressing it here.

\section{Dataset}
\label{section:dataset}

The Visible Human Project (VHP) \cite{VisibleHuman} provides a publicly available dataset that contains cross-sectional cryosection, CT scans, and MRI images from one male cadaver and one female cadaver. In our work, we take 33 MRI samples of head, 328 CT scans of thorax and abdomen, and 696 color cryosections from Visible Human Male to perform color transfer on the monochromatic MRI images. Among the color cryosection images, 40 of them are from male-head, and remaining 656 images are from male-thorax and abdomen. Here, the monochromatic images are our target images, and all of the color cryosection slices form the database of reference images. After color transfer, we perform ray casting for medical volume rendering where we use the stack of colored medical images as our volumetric data. 

\begin{figure*}[t]
  \centering
  \includegraphics[width=0.8\textwidth]{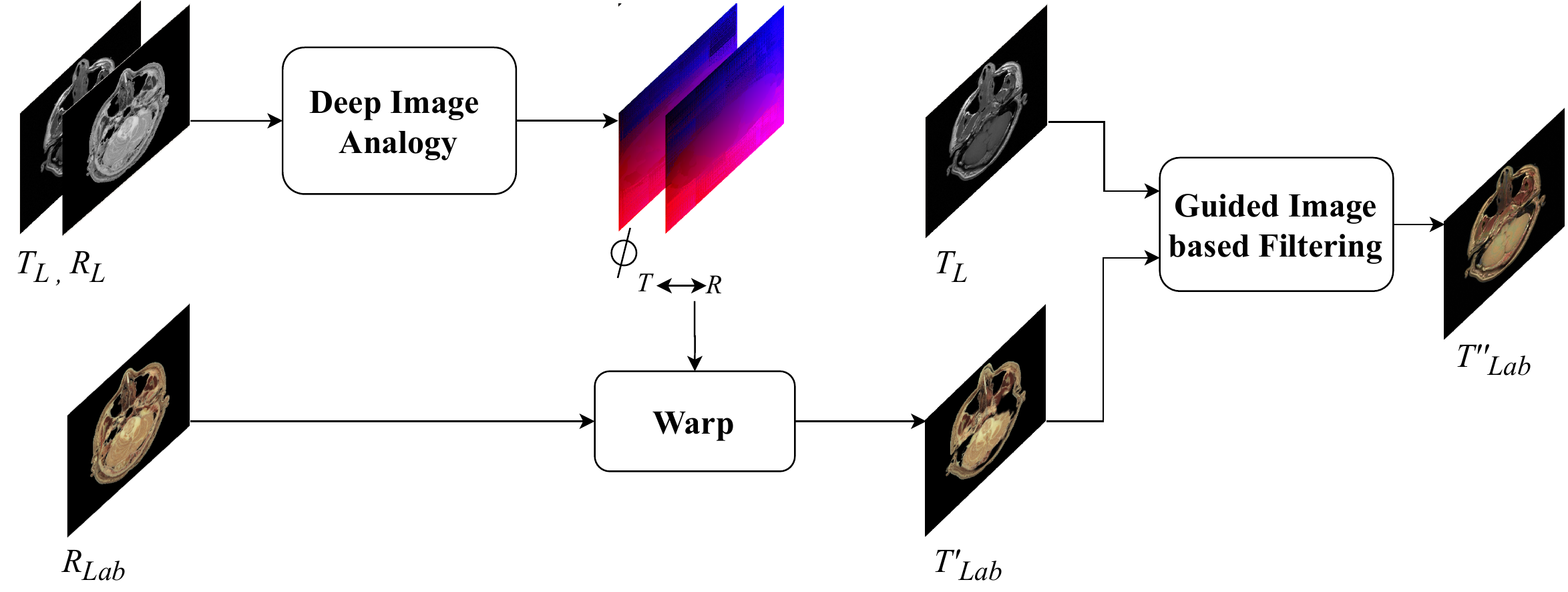}
  \caption{Our proposed system pipeline consists of two stages. In the first stage, we perform deep image analogy on two input images $T_L$ and $R_{Lab}$ to obtain the reconstructed output $T’_{Lab}$. Then, in the second stage, we use guided image based filtering technique to perform colorization of image $T_L$ with the help of reconstructed output $T’_{Lab}$ }
  \label{fig:SystemPipeline}
  \Description{Insert Description here}
\end{figure*}

\begin{figure}[ht]
\Description{Caption 1}
\includegraphics[width=1\linewidth]{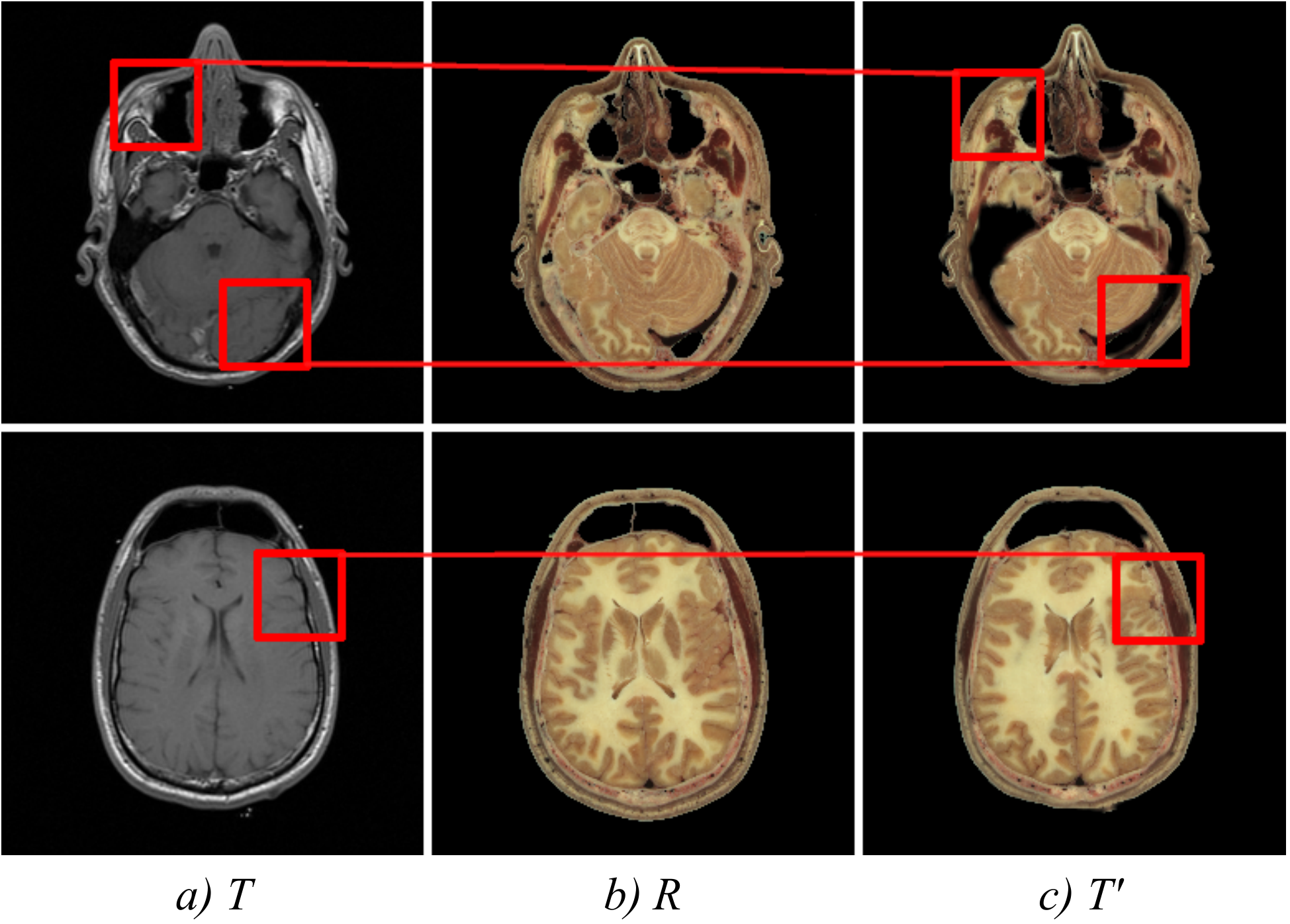} 
\caption{Images in first column are two MRI slices of Visible Human\cite{VisibleHuman} Head  which represent the target images $T$, images in second column are the cross-sectional cryosections of Visible Human Head which represent the reference images $R$, and images in third column  are the reconstructed output $T'$ obtained after performing visual attribute transfer with Deep Image Analogy on images in first and second column. We can see unwanted distortions in the output image that are not present in the target images. (Three such distortions are indicated by red squares connected with straight lines)
}
\label{fig:deepImageAnalogyOutput}
\end{figure}

\begin{figure*}[thbp]
    \Description{Caption 1}
    \includegraphics[width=0.74\textwidth]{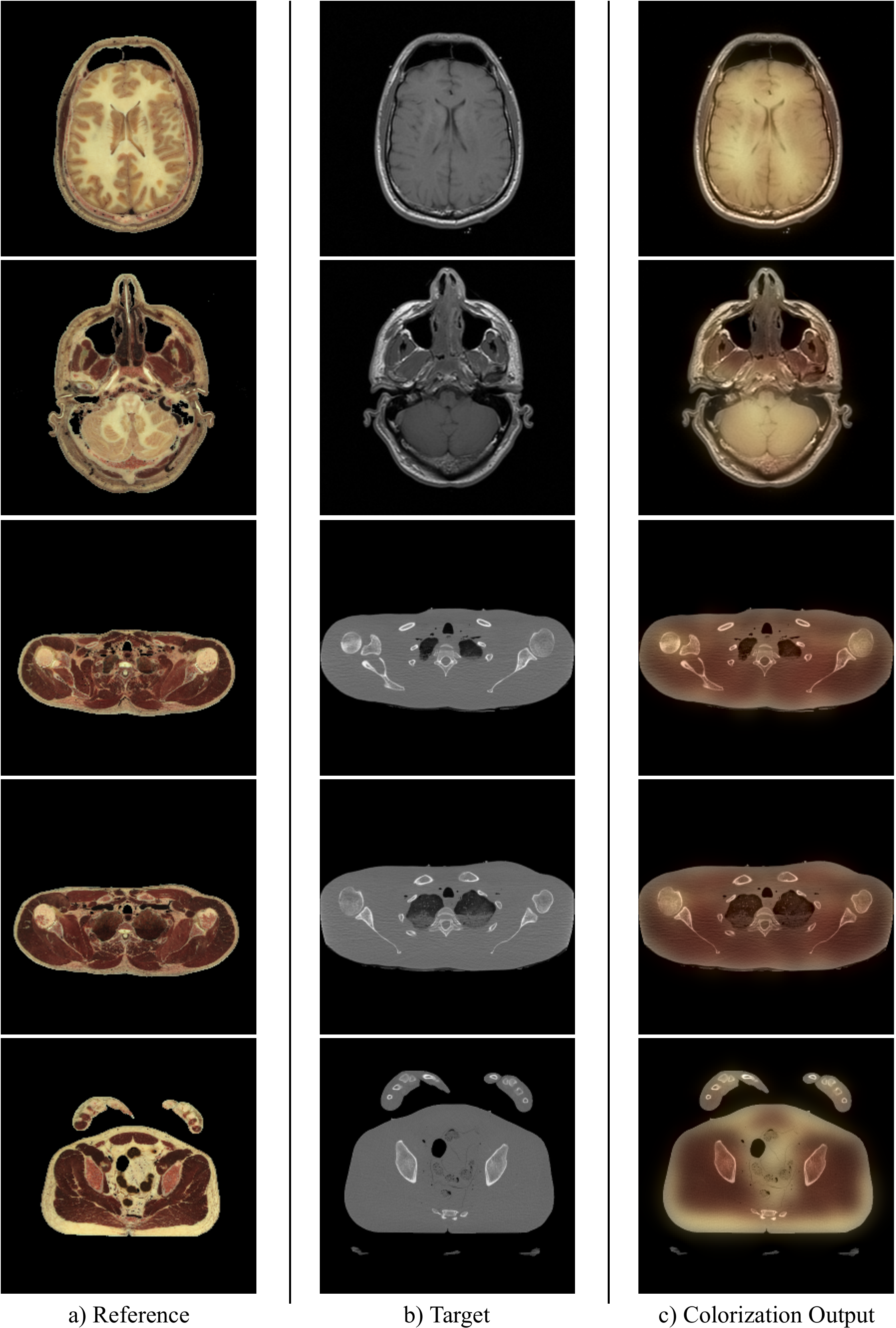} 
    \caption{ With the use of Fast Global Smoothing filter\cite{fgsfilter}, we obtain the final colorized version (third column) of the target grayscale medical images presented in the second column with the reference images presented in the first column. Target images (from top to bottom): two MRI scans of head, two CT-scans of thorax, and one CT-scan of abdomen. Reference Images (from top to bottom): two cross-sectional cryosections of head, two of thorax and one of abdomen from. Both of the target and reference images are from the Visible Human dataset\cite{VisibleHuman}.
    }
    \label{fig:colorization-methodology}
\end{figure*}

\section{Methodology}
\label{section:methodology}

Our end goal is to automatically transfer color or tone to a stack of monochromatic medical images from semantically similar reference images and to finally render a 3D volume using the colored image stack. Therefore, there are two parts to our problem: automatic reference based color transfer and medical volume rendering. 

In our approach to tackle the first problem, we use a state of the art dense correspondence algorithm like Deep Image Analogy \cite{Liao17} which finds a semantically meaningful dense correspondence between two input images. The reason behind establishing a dense correspondence is to perform color transfer between semantically similar regions of reference image and target image. Since, the reference images that we are using here are frozen cross sections of the human body, we can argue that the range of colors present in the reference image will be sufficient for colorization of target images.

\subsection{ Deep Image Analogy}
\label{section:meth-deepImageAnalogy}

Using Deep Image Analogy \cite{Liao17}, given that there are two input images $T$ and $R$, we can construct an image $T'$, which is semantically similar to $T$ but contains the visual attributes of $R$. $T’$ is the reconstructed result after transfer of visual attributes like color, texture, and style. Liao et al. performed two major steps in their work for dense matching. In the first step, their implementation computes features of two input images $T$ and $R$ at different intermediate layers of a pretrained CNN that is trained on image recognition tasks. In the second step, they use the PatchMatch algorithm \cite{Barnes10} in deep feature domain rather than in image domain to find deep correspondence between those images. The final output after these two steps is a bidirectional similarity mapping function
$ \phi_ {T \leftrightarrow R} $
between those two images. PatchMatch is a randomized correspondence algorithm which finds an approximate nearest neighbor field (NNF) between two images. The algorithm has three main components: \textbf{initialization}, \textbf{propagation} and \textbf{random search}, where it first initializes NNF as a function
$ f:\mathbb{R}^2 \rightarrow \mathbb{R}^2 $
of random offsets. This is followed by an iterative process, where good patch offsets are propagated to adjacent pixels and and finally, a random search is performed in the neighborhood of the best offset.

The pretrained network used by Liao et al. \cite{Liao17} to obtain feature representations of an image is the original VGG-19 \cite{VGG19} network that has been pre-trained on the colored ImageNet dataset \cite{ImageNet}. However, finding good semantic similarity between two images is difficult when one of them is grayscale and the other is color. Furthermore, original VGG-19 has a comparatively poor performance when it comes to image recognition on grayscale images \cite{He18}. So, in order to minimize the difference in performance, we use a gray-VGG-19 as the feature extractor network. Gray-VGG19 was trained only using the luminance channel of images in the ImageNet dataset and the performance evaluation of gray-VGG-19 shows that it exhibits superior performance to original VGG-19 when both of these networks are tested on gray images. Original VGG-19 tested on gray images achieves 83.63\%, while gray-VGG-19 achieves 89.39\% on the Top-5 Class accuracy\cite{He18}.

The system pipeline to perform automatic color transfer is shown in figure \ref{fig:SystemPipeline}. We propose our system with CIE Lab color space which is perceptually linear and it expresses color as three values $L$, $a$, and $b$. Here, $L$ denotes Luminance, and $a$ and $b$ denote Chrominance values. Thus each image can be divided into three channels: a single luminance channel and two chrominance channels. Our system takes two images: a grayscale target image and a color reference image, which can be denoted as  
$ T_L \in \mathbb{R}^{h*w*1} $  
and 
$ R_{Lab} \in \mathbb{R}^{h*w*3} $ 
respectively. Here, $h$ and $w$ are the height and width of the input images. The output from deep image analogy is the bidirectional mapping function between the two images which is a spatial warping function defined with bidirectional correspondences. The bidirectional mapping functions can be denoted as 
$ \phi_ {T \leftrightarrow R} $, which returns the transformed pixel location for a given source location $'p'$. It consists of a forward mapping function 
$ \phi_ {T \rightarrow R} $, that maps pixels from T to R, and a reverse mapping function $ \phi_ {R \rightarrow T} $, that maps pixels from R to T.  Using the mapping function $ \phi_ {R \rightarrow T} $, we map the image $R_{Lab}$ to image $T’_{Lab}$. In other words, $T’_{Lab}$ is the reconstructed result obtained after warping the image $R_{Lab}$ with the mapping function $ \phi_ {R \rightarrow T} $.

Figure \ref{fig:deepImageAnalogyOutput}c) shows the reconstructed output images that were obtained after performing deep image analogy on the given target and reference images in figure \ref{fig:deepImageAnalogyOutput}a) and figure \ref{fig:deepImageAnalogyOutput}b). We can observe that the reconstructed result $T’_{Lab}$ not only transferred color but other attributes like style and texture that were extracted from the reference image $R_{Lab}$. The results also contain distortions that are unwanted in our final colored image.  In a field like medical imaging, edges and corners are key features and the overall structural details must be preserved when performing any kind of image manipulation.

In our approach to tackle this problem, where we need to maintain the semantic structure from the target image $T_L$, preserve color, and disregard other visual details from the reference image, we performed guided image based filtering technique to obtain the final colored image $T''_{Lab}$. Guided image based edge preserving filters like Domain-Transform (DT) filter \cite{dtfilter}, the Guided Filter (GF) \cite{gifilter}, the Weighted Least Square Filter (WLS) \cite{wlsfilter} and the Fast Global Smoother (FGS) filter \cite{fgsfilter} are designed to perform smoothing of an image while preserving the edges using the content of the guide image. Specifically, we use the Fast Global Smoother\cite{fgsfilter} to smooth both input image $T_L$ and $T'_{Lab}$ with the guidance $T_L$. More on this is explained in the following section.

\subsection{Colorization using Guided Image Filters} 
\label{section:meth-colorization}

As seen in figure \ref{fig:deepImageAnalogyOutput}, the reconstructed output after performing Deep Image Analogy contains unwanted structural distortions that were introduced from the reference image. Since our goal is to only transfer color from the reference image, we resorted to using guided image based edge preserving smoothing filters to preserve the color and structural details of the target image while removing other visual artifacts from image $T'_{Lab}$. One such filter is the Fast Global Smoother \cite{fgsfilter}, which performs spatially inhomogeneous edge-preserving smoothing. To be more specific, we performed
\begin{equation}\label{eq:colorization}
T''_{Lab} = fgs( T'_{Lab} , T_L ) - fgs( T_{Lab} , T_L ) + T_{Lab}
\end{equation}
where function $fgs(.)$ , which is a short-hand notation for the Fast Global Smoothing filter, takes two input images: a source image, and a guide image whose edges guide the smoothing operation for source image. First we perform the edge aware smoothing operation on images $T’_{Lab}$ and $T_{Lab}$ with guide image $T_L$ to get smoothened output $ fgs( T'_{Lab} , T_L ) $ and $ fgs( T_{Lab} , T_L ) $, then we add the differences between these two filtered output to image $T_{Lab}$ to get the final colored image $T”_{Lab}$. Here, since the target image $T_{Lab}$ is a grayscale image, its chromatic color channels $T_a$ and $T_b$ contain true neutral gray values, i.e zero values.  

Figure \ref{fig:colorization-methodology}c) shows the colorization result obtained from equation \ref{eq:colorization}. We can observe that the final colorized version of the target images are free from undesirable distortions and attributes that were previously being introduced from the reference image.

\begin{figure}[t]
    \Description{Caption 1}
    \includegraphics[width=1\linewidth]{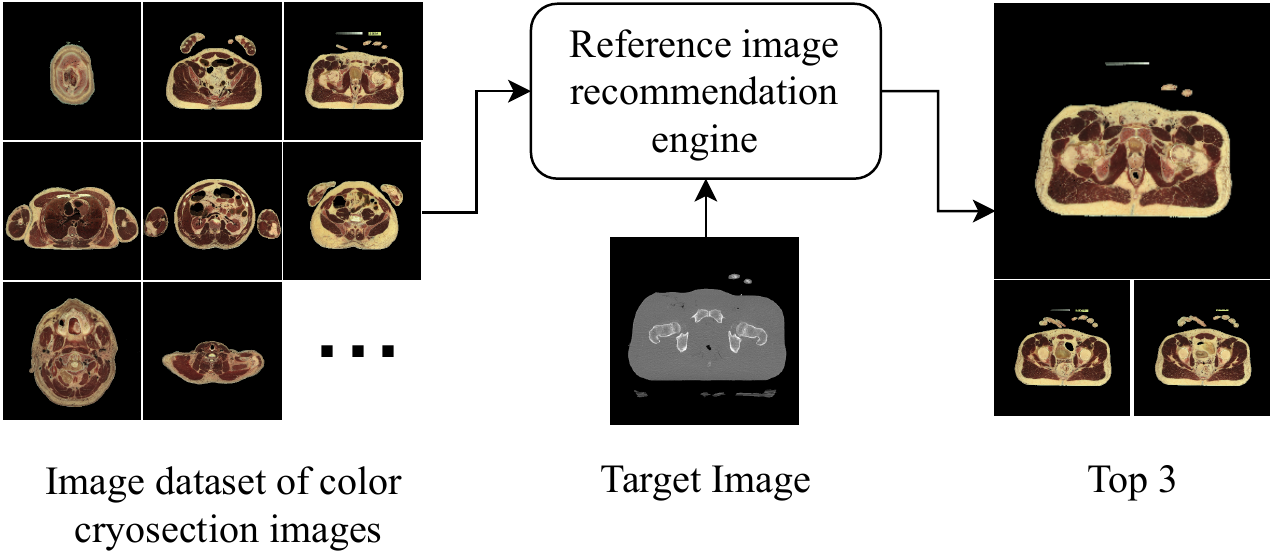} 
    \caption{ Pipeline of reference image retrieval system, where the recommendation engine takes two inputs: a monochromatic target image, and 696 color cryosection images from Visible Human head, thorax and abdomen. The output of the recommendation engine are the top three reference images which share similar semantic content with the target image. 
    }
    \label{fig:recommendationEngine}
\end{figure}

\begin{figure*}[ht]
  \centering
  \includegraphics[width=\textwidth]{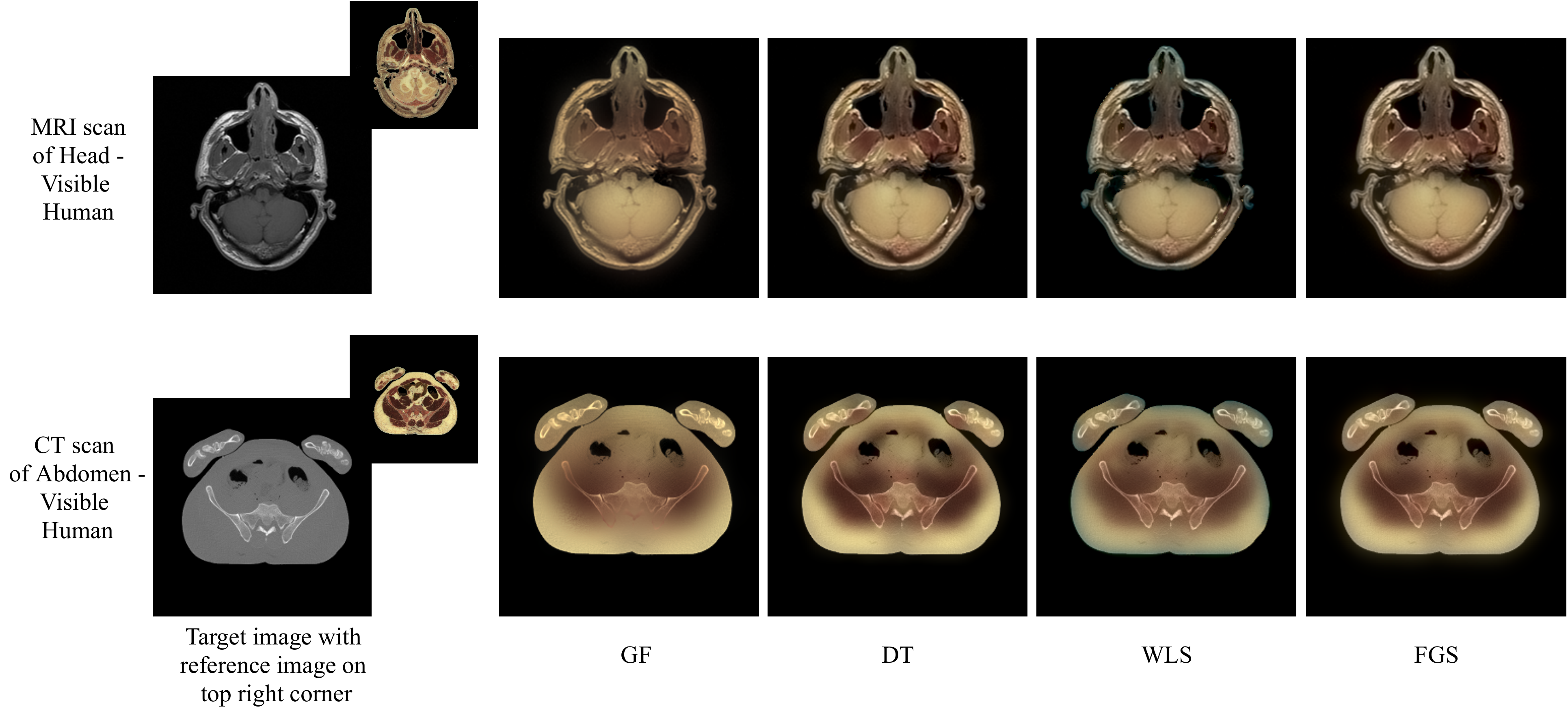}
      \caption{ Comparison of results obtained after performing Deep Image Analogy followed by Colorization using guided image filters. 
    First column: Target images with reference image on the top right corner of each target image. 
    Second column: Colorized result with Guided Filter\cite{gifilter}, where radius $r= 16$, regularization term $\epsilon=2$
    Third column: Colorized result with Domain Transform Filter\cite{dtfilter}, where sigma spatial $\sigma_s=8$, sigma range(color) $\sigma_r=200$
    Fourth column: Colorized result with Weighted Least Square filter\cite{wlsfilter}, where lambda $\lambda=0.2$, alpha $\alpha=1.8$
    Fifth column: Colorized result with Fast Global Smoother filter \cite{fgsfilter}, where lambda $\lambda=32$, sigma range $\sigma_r=200$.
    Note that the reconstructed output obtained after performing Deep Image Analogy is not shown in the figure above.}
  \label{fig:filterComparision}
  \Description{Insert Description here}
\end{figure*}

\begin{figure*}[ht]
  \centering
    \includegraphics[width=1\textwidth]{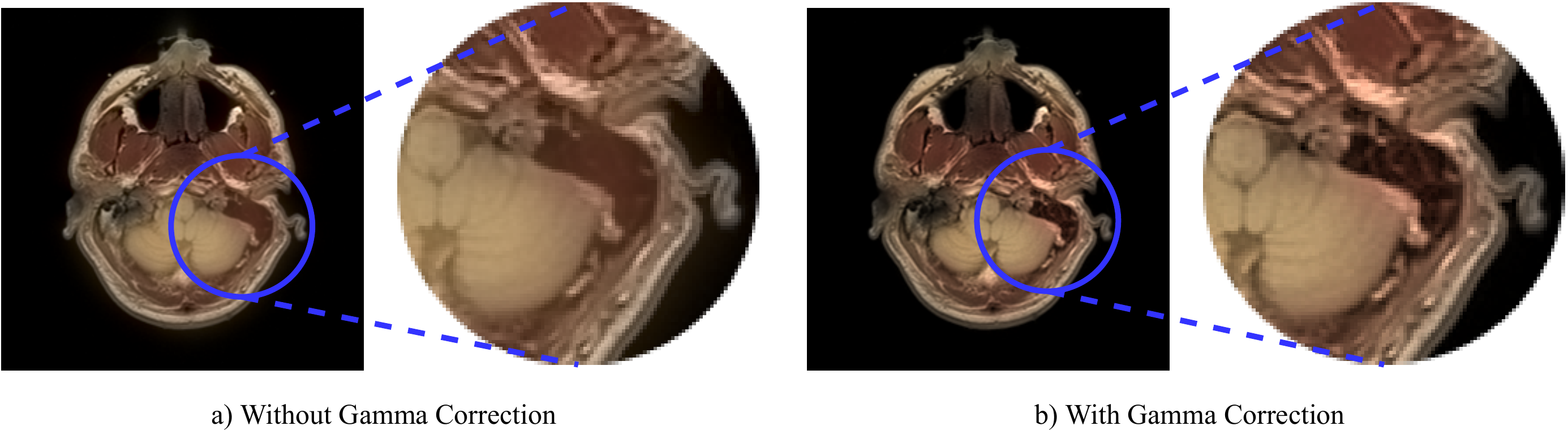} 
    \caption{ Comparison of results obtained $a)$ without gamma correction and $b)$ with gamma correction. The circular image in the right-side of each image shows a closer view of the region present inside the blue circle. With gamma correction, the colored image on the bottom appears more detailed and at the same time color spilling out of the edges is also reduced.
    }
    \label{fig:gammaCorr}
\end{figure*}

\subsection{Reference Image Retrieval}
\label{section:meth-referenceImageRetrieval}

To aid for selection of good reference images, we propose to use a robust image retrieval algorithm that automatically recommends the best reference images for performing color transfer. Figure \ref{fig:recommendationEngine} shows the pipeline of the reference image recommendation system. The inputs to the recommendation engine are the dataset of our reference images which contain 696 frozen cross-section slices of Visible Human Male and a single monochromatic query image. Ideally, the image suggested by the recommendation engine is expected to have similar semantic structure as the query image.

There have been several works regarding content based image retrieval with the use of deep neural networks \cite{Babenko14}\cite{Wan14}\cite{Ng15}\cite{Razavian14}. Content based Image Retrieval techniques that use deep neural networks exploit the property of pretrained CNN networks, where the lower layers (layers closer to the input) of the network encode low-level details of an image and the higher layers are more sensitive towards high-level details, that is, content of an image. Since the features obtained from the higher layers provide a good representation of the content of an image, it is common to use these features, which are also called descriptors, to evaluate semantic similarity between two images.

A similar work on image retrieval has been performed by \cite{He18} where they obtain features of reference images and a query image from two different layers of gray-VGG-19 network and finally compute a similarity score to rank the images. We adopt a similar technique, where we first feed all of our reference images to VGG-19 network and pre-compute their feature vector from the first fully connected layer $F^6_c$. In the second step we feed our target image through the network to obtain its features from the same fully connected layer of VGG-19 network.

Let $f_{R_i}$ denote the feature vector of $i^{th}$ reference image and , $f_T$ denote the feature vector of the target image. To obtain the measure of semantic similarity between two images, we compute cosine similarity, as given by equation \ref{eq:cos_sim}, between these feature vectors for each reference-target image pair
${R_i , T} $. We use this similarity score to obtain the best reference image for performing color transfer. 
\begin{equation}\label{eq:cos_sim}
d(f_{R_i}, f_t) = \frac{f_{R_i} . f_t}{|| f_{R_i} || \times || f_t ||}
\end{equation}

\begin{figure*}[ht]
  \centering
    \includegraphics[width=0.9\textwidth]{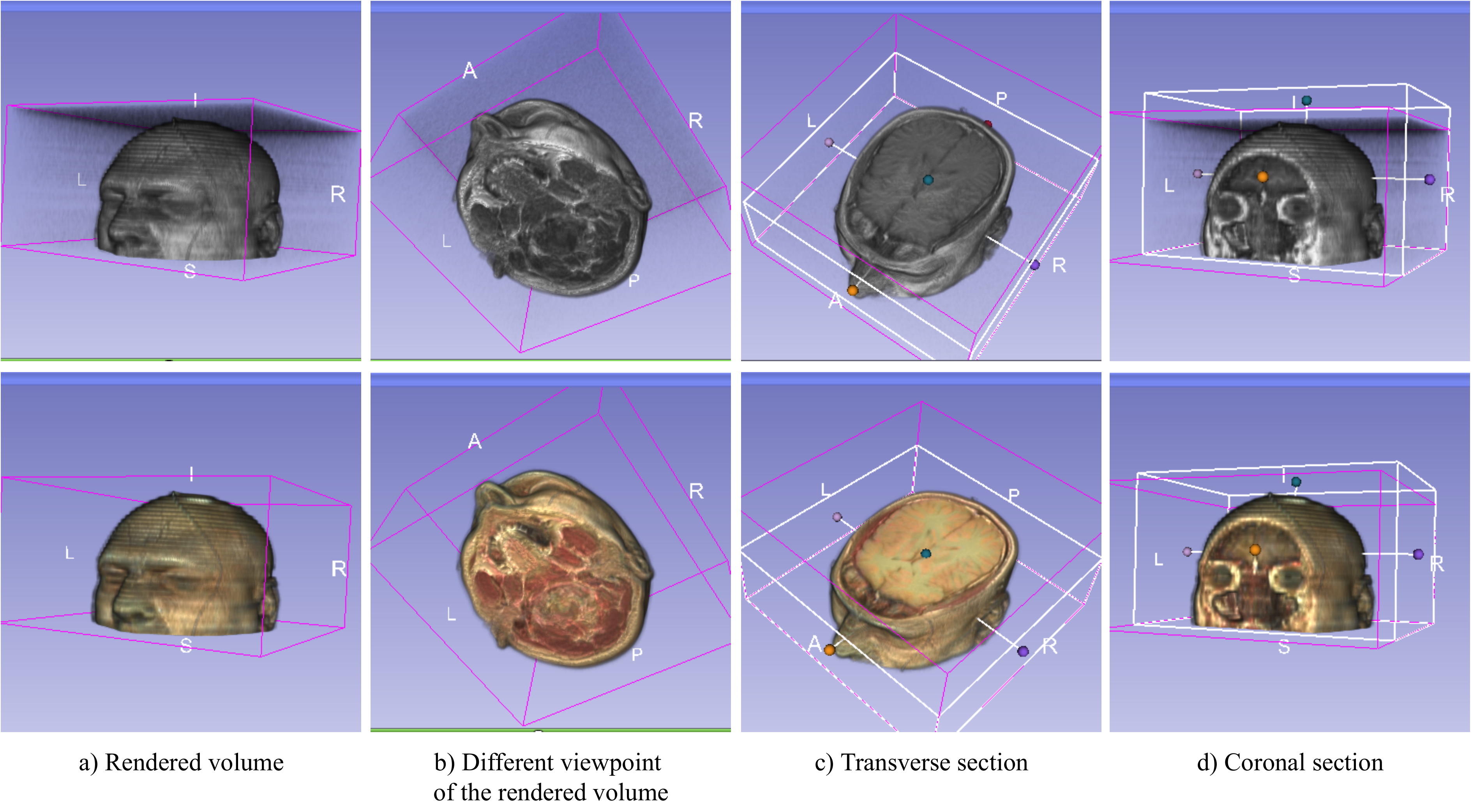} 
    \caption{ Volume visualization using 3D slicer for top half of Visible Human Head. First row: Volume is rendered with 33 monochromatic MRI images. Second row: Volume is rendered with corresponding colored versions of medical images that are used to render the volume displayed in the first row.
    }
    \label{fig:volVisHead}
\end{figure*}

\begin{figure*}[ht]
  \centering
    \includegraphics[width=0.9\textwidth]{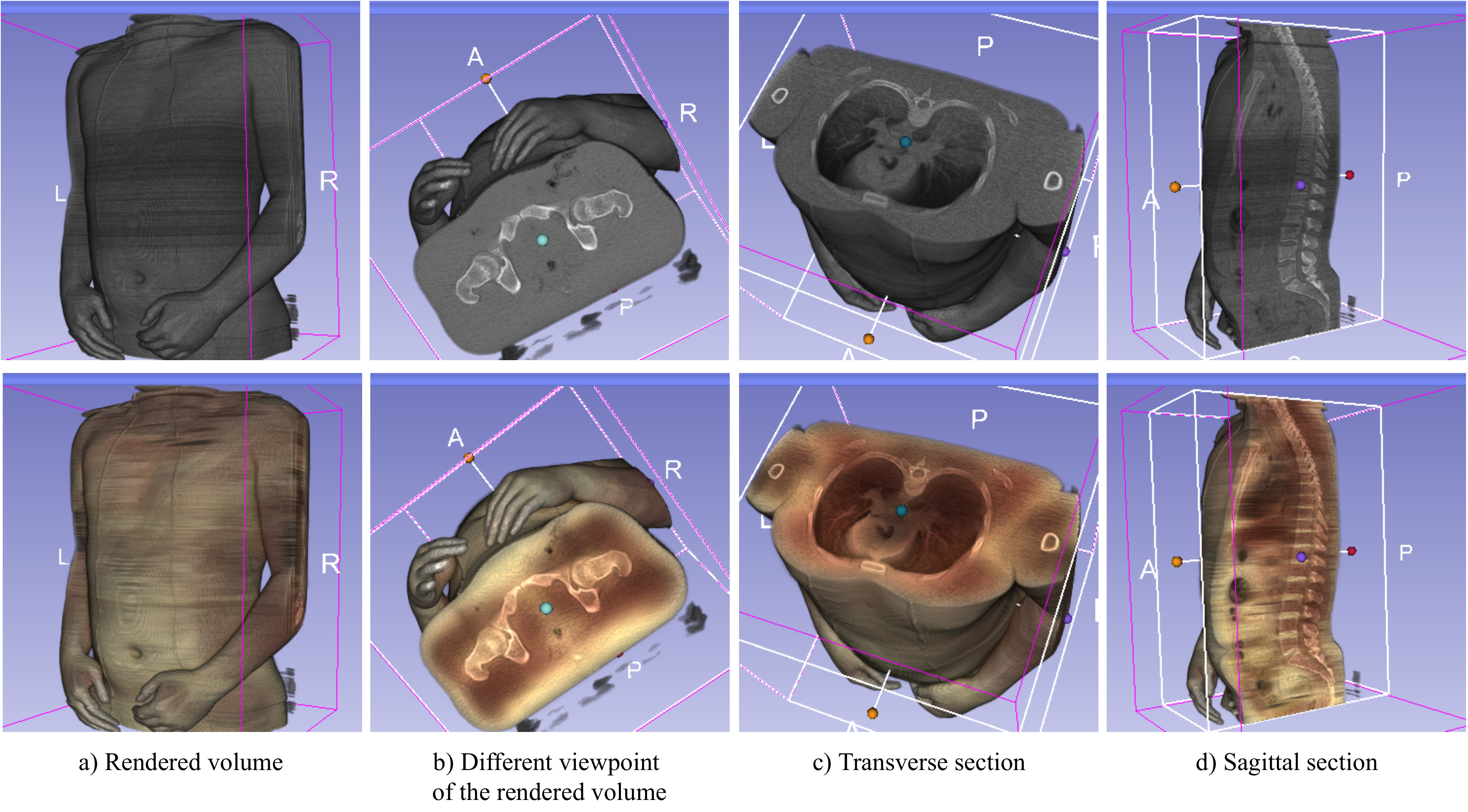} 
    \caption{ Volume visualization using 3D slicer for Visible Human thorax and abdomen. First row: volume is rendered using 328 monochromatic CT scans. Second Row: Volume is rendered with corresponding colored versions of medical images that are used to render the volume displayed in the first row.
    }
    \label{fig:volVisAbdomen}
\end{figure*}

\subsection{Direct Volume Rendering}
\label{section:meth-DVR}

After obtaining the colored images, we perform volume visualization using 3D-slicer \cite{3Dslicer}, which is an open source medical image computing platform. 3D-slicer provides various tools for biomedical research and GPU-based Direct Volume Rendering is one of them. Typically, the final volume is rendered using a stack of medical images, where ray-casting is performed on the volumetric medical data. The ray-casting module is freely available on 3D slicer and has been used in many studies \cite{Li10}\cite{Zhang10}\cite{Schubert11}\cite{Meyer09}.

\section{Results}
\label{section:results}

As mentioned in section \ref{section:meth-colorization}, to remove the unwanted artifacts introduced from reference image to target image, we performed guided image filtering on the reconstructed output from Deep Image Analogy. Figure \ref{fig:filterComparision} shows the comparison of results that were obtained after performing colorization with different guided image filters. We obtain the output in the same way as mentioned in equation \ref{eq:colorization} where the fast global smoothing filter (FGS) is replaced by other filters: Guided filter (GF), Domain Transform filter (DT),and Weighted Least Square filter (WLS). A comparison between the four filters provided by Min et al.\cite{fgsfilter} shows that fgs provides the speed advantage over the other three filters, and unlike GF and DT filters, it is not prone to halo effect. Due to these benefits, the FGS filter is the preferred filter in our colorization pipeline.

\subsection{Gamma Correction}
\label{section:results-gammaCorrection}

With color transfer, in addition to ensuring that all the semantic structures present in the grayscale image are preserved in the final colored image, we also aim to have our colorized result look as natural as possible. In some of the resulting images that we obtained after performing color transfer, the image seemed to lack proper contrast making the image appear to have a washed-out effect. Furthermore, it was noticed that, in some regions of those images, color was spilling out of the edges. This may happen due to excessive smoothing of image with guided image filters. An example of this can be seen in figure \ref{fig:gammaCorr}. So, to improve the vividness of the result, we use gamma correction which is simple non-linear operation defined by the power law transformation as given in equation \ref{eq:gammaCorr}.
\begin{equation} \label{eq:gammaCorr}
    I_{in} = \alpha I^{\gamma}_{out}
\end{equation}

Specifically, after obtaining the reconstructed image from deep image analogy, we perform gamma encoding on both target and the reconstructed image, where we use a gamma value of $\gamma=1/2$. This is followed by colorization using guided image filters, and finally we perform gamma decoding with gamma value of $\gamma=2$ to obtain the resulting colored image. With gamma correction, not only the color content spilling out of the edges was suppressed, but also the vividness of the image was improved to a certain extent, making the image look comparatively more sharp and detailed.

\subsection{Volume Visualization}
\label{section:results-VolVis}

Since there is no ground-truth for this task, we only present qualitative results here that we obtained in our work. After obtaining the colored version of all the monochromatic medical images, we perform volume visualization using 3D slicer \cite{3Dslicer}. Specifically, we use the GPU based ray-casting module provided in 3D slicer for volume visualization. First row in Figure \ref{fig:volVisHead} and \ref{fig:volVisAbdomen} show the volume rendered using a stack of grayscale medical images while the second row in those two figures show the volume rendered using colored medical images. 

Here, we present the rendering result with two different image sets. In figure \ref{fig:volVisHead}, the first row displays the rendered volume for the top half of the human head, where we use 33 MRI scans obtained from the Visible Human Dataset. Similarly, the first row of figure \ref{fig:volVisAbdomen} displays the rendered volume for Visible Human thorax and abdomen, where we use 328 monochromatic CT scans. In the second row of figure \ref{fig:volVisHead} and \ref{fig:volVisAbdomen}, we present a similar viewpoint of the volume rendered in the first row, but in this case, the volume visualization is rather performed with colored versions of MRI images and CT scans.

To perform visual inspection on the quality of the rendered volume we also compare the transverse section, coronal section and sagittal section between grayscale volume and colored volume as shown in figure \ref{fig:volVisHead}c) , \ref{fig:volVisHead}d), \ref{fig:volVisAbdomen}c) and \ref{fig:volVisAbdomen}d). We can see that the structure of grayscale volume is well preserved in the final colored volume, but at the same time, we also notice the presence of some dark patches in the colored volume, figure \ref{fig:volVisAbdomen}b), c) and d). On inspection of the grayscale viewpoint of the same volume, we can tell that these patches are not desired in the final colored volume and they were most definitely introduced because of unwanted structures present in the reference images.

\section{Conclusion}
\label{section:conclusion}

In our work, we have managed to introduce a new pipeline for medical volume visualization. First, we explored the use of visual attribute transfer with deep image analogy, and then with the introduction of guided image based filtering technique, we were able to perform automatic colorization of grayscale medical images. After that, we extended our work for direct volume rendering, where we make use of the ray-casting module provided by 3D slicer to perform volume visualization using colored medical images. Furthermore, we also introduced a robust reference image retrieval system which aims to recommend the best reference image in terms of semantic similarity with the target image to perform color transfer. 

With our proposed pipeline, we completely eliminate the tedious work of interacting with a volume renderer where a user has to experiment with different values on a transfer function to generate plausible color in the rendered volume. We also believe that our approach will be very helpful for a variety of applications in the medical community and prove to be more useful in performing anatomical analysis during clinical procedures.

\section{Discussion}
\label{section:discussion}

In our work, we are successfully able to extend color transfer with deep image analogy for medical volume visualization. The core of our work is based on the work performed by Liao et al. \cite{Liao17} and since visual attribute transfer with deep image analogy has its own drawbacks , we implicitly inherit all of those limitations in our work.

Transfer of color or other attributes by finding out semantic similarities between two images is not always reliable. The overall success of our work also relies on the performance of a pretrained VGG-19 network. Our system pipeline starts with extracting features from the images and the network itself may not always be successful in extracting feature representation of all the content present in the image. An alternative approach that we can take here is to experiment with other pre-trained networks that have superior performance than VGG-19. Then we can retrain the network on a medical image dataset before using it as a feature extractor network. 

Furthermore, when the reference image has a significant amount of semantic differences with the target image, the system fails to find good correspondence and as a result we see unsmooth colored artifacts in the final images. Examples of that are seen in the second row of figure \ref{fig:filterComparision}, where the resulting colored CT scans of Visible Human Abdomen contain unwanted dark patches which were transferred from the reference images. In cases like these, it is difficult to perform reliable color transfer from one image to another.

\begin{acks}
We thank anonymous reviewers for their valuable feedback. We also acknowledge the authors of \cite{Liao17} and \cite{wlsfilter} for making their work publicly available. 
\end{acks}

\bibliographystyle{ACM-Reference-Format}
\bibliography{references}


\end{document}